\pdfoutput=1

\documentclass[11pt]{article}

\usepackage[final]{acl}

\usepackage{times}
\usepackage{latexsym}

\usepackage[T1]{fontenc}

\usepackage[utf8]{inputenc}

\usepackage{microtype}

\usepackage{inconsolata}

\usepackage{graphicx}
\usepackage{amsmath}
\usepackage{amssymb}
\usepackage{amsthm}
\usepackage{enumitem}
\usepackage{changepage}
\theoremstyle{definition}
\usepackage{booktabs}
\usepackage{graphicx}
\usepackage{subcaption}
\usepackage{multirow}
\definecolor{lightgray}{gray}{0.92}
\usepackage{listings}
\usepackage[most]{tcolorbox}
\usepackage{bbm}
\newtheorem{definition}{Definition}

\title{DEL-ToM: Inference-Time Scaling for Theory-of-Mind Reasoning via Dynamic Epistemic Logic}

\author{
\\
\textbf{Yuheng Wu}\textsuperscript{1} \hspace{2em} 
\textbf{Jianwen Xie}\textsuperscript{2} \hspace{2em} 
\textbf{Denghui Zhang}\textsuperscript{3,\textdagger} \hspace{2em}
\textbf{Zhaozhuo Xu}\textsuperscript{3,\textdagger} \\
\textsuperscript{1}Stanford University \hspace{1em} 
\textsuperscript{2}Lambda, Inc. \hspace{1em} 
\textsuperscript{3}Stevens Institute of Technology\\
\texttt{yuhengwu@stanford.edu} \hspace{1em} \texttt{jianwen.xie@lambda.ai} \hspace{1em} \texttt{\{dzhang42,zxu79\}@stevens.edu}\\
}

\begin{document}

\maketitle

\begingroup
  \renewcommand{\thefootnote}{\textdagger}
  \footnotetext{Corresponding authors: \{zxu79,dzhang42\}@stevens.edu.}
\endgroup

\begin{abstract}
Theory-of-Mind (ToM) tasks pose a unique challenge for large language models (LLMs), which often lack the capability for dynamic logical reasoning. In this work, we propose DEL-ToM, a framework that improves verifiable ToM reasoning through inference-time scaling rather than architectural changes. Our approach decomposes ToM tasks into a sequence of belief updates grounded in Dynamic Epistemic Logic (DEL), enabling structured and verifiable dynamic logical reasoning. We use data generated automatically via a DEL simulator to train a verifier, which we call the Process Belief Model (PBM), to score each belief update step. During inference, the PBM evaluates candidate belief traces from the LLM and selects the highest-scoring one. This allows LLMs to allocate extra inference-time compute to yield more transparent reasoning. Experiments across model scales and benchmarks show that DEL-ToM consistently improves performance, demonstrating that verifiable belief supervision significantly enhances LLMs' ToM capabilities without retraining. Code is available at \href{https://github.com/joel-wu/DEL-ToM}{https://github.com/joel-wu/DEL-ToM}.

\end{abstract}

\section{Introduction}
\begin{figure*}[t]
    \centering
    \includegraphics[width=0.95\linewidth]{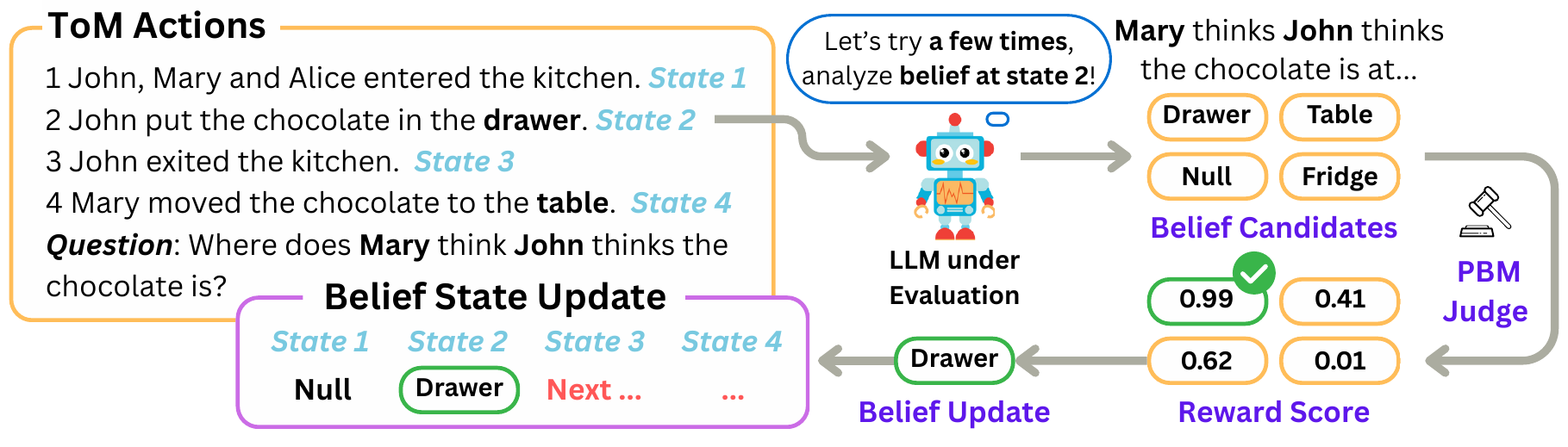}
    \caption{Overview of the DEL-ToM framework. Each belief state is inferred from the previous state and the current action. The LLM generates multiple candidate belief traces in parallel, and the PBM assigns reward scores to filter a top-scoring subset, which is then used to continue reasoning toward the next belief state.}
    \label{fig:intro}
\end{figure*}

\begin{quote}
``To know what John knows is to know the worlds that are compatible with his belief, and to know which ones are not.''
\hfill \hfill --- \textit{Jaakko Hintikka}~\cite{Hintikka1989-HINTLO-7}
\end{quote}

\noindent The ability to attribute beliefs, desires, and intentions to others, known as Theory-of-Mind (ToM)~\cite{premack1978does, cdennett_1978_beliefs, apperly2009humans}, is a fundamental component of social intelligence~\cite{baron1991precursors}. ToM enables agents to reason about what others think, want, or know, and to anticipate their subsequent behavior~\cite{rabinowitz2018machine}. 

Recent studies suggest that large language models (LLMs)~\cite{brown2020language} exhibit ToM abilities~\cite{strachan2024testing, lin2024constrained, street2024llms, amirizaniani2024llms, sclarexplore, wu2025large}. However, ToM performance follows a scaling law~\cite{kosinski2023theory}, with smaller models showing limited ability on ToM tasks. This limitation poses a challenge for low-resource deployments, where edge agents are expected to robustly infer users’ intentions and act in alignment with human expectations. At the same time, current evaluations compare only the final output to the ground-truth label~\cite{chen-etal-2024-tombench}, leaving it unclear whether correct answers result from genuine reasoning or from lucky guessing~\cite{ullman2023large}. Consequently, existing ToM reasoning remains unverifiable and not applicable in practice. This paper addresses the question: \textit{How can we enable LLMs to perform verifiable ToM reasoning, especially in low-resource settings?}

Following process reliabilism~\cite{goldman_1979_what}, verifiable ToM reasoning requires a sequence of intermediate belief states that reliably support the final conclusion. We formalize this reasoning process using Dynamic Epistemic Logic (DEL)~\cite{baltag98logic, van2001games, plaza2007logics, van2007dynamic, aucher2013complexity}, a logic system grounded in the traditions of formal logic and semantics~\cite{Frege1879-FREBAF-2, Russell1910-RUSPMV, Wittgenstein1922-WITTL-11, Tarski1956-TARTCO, Hintikka1962-HINKAB-3, Kripke1963-KRISCO}. DEL models agents’ beliefs with epistemic models, actions with event models, and belief change via product updates, allowing us to view ToM reasoning as dynamic logical reasoning.

Within this framework, transparent belief traces are generated and evaluated by a Process Belief Model (PBM). By scoring multiple candidates, the PBM enables us to select the most reliable trace. This constitutes inference-time scaling: spending more computation during inference to obtain more reliable reasoning traces, which in turn allows smaller models to achieve stronger ToM performance while remaining efficient for deployment. We experiment with different trace selection and search strategies for ToM reasoning.

To train the PBM, we first generate ToM-related questions and use DEL to produce belief process labels. We then use GPT-4o-mini~\cite{hurst2024gpt} to answer these questions. Finally, DEL-generated gold labels are used to automatically score GPT-generated traces, producing positive and negative examples for PBM training. Unlike other process-level reward modeling datasets, which rely on human annotation or LLM assistance~\cite{wang2024math}, our labels are derived from a formal DEL system, which guarantees correctness.

In conclusion, we approach ToM reasoning through the lens of formal logic. Using a PBM trained via DEL, we make each intermediate belief update explicit and employ search-based methods to select the most reliable trace. This enables inference-time scaling and yields dynamic logical reasoning grounded not only in model outputs, but in verifiable, structured belief updates. Our contributions are threefold:

\begin{itemize}[nosep,leftmargin=*]
    \item We propose a new perspective on ToM reasoning by framing it as a problem of process reliability. By modeling reasoning as a multi-step dynamic belief-update process, we can apply inference-time scaling to select more reliable belief traces.
    
    \item We formalize ToM reasoning in the framework of DEL and construct a PBM dataset with noise-free supervision derived from DEL. This enables training PBMs for stepwise reasoning evaluation.
    
    \item We evaluate our approach across different model scales and search strategies. Our method consistently improves LLM performance on standard ToM benchmarks.
\end{itemize}

\section{Background and Motivation}
\noindent \textbf{ToM in LLMs.} 
Researchers have designed various tasks to evaluate the ToM capabilities of LLMs. Among these, false belief tasks are the most widely used, typically in two forms:

\begin{itemize}[nosep,leftmargin=*]
\item Unexpected Contents: A protagonist is shown an object with misleading external cues (e.g., an opaque crayon box that actually contains candles). The LLM under evaluation must identify that the actual content is candles while recognizing that the protagonist holds the mistaken belief that the box contains crayons.
\item Unexpected Transfer: An object is moved without the protagonist’s knowledge, and the LLM must predict where the protagonist will search for it -- based on the protagonist’s outdated belief.
\end{itemize}

Among the two, the unexpected transfer task is more commonly used. Figure~\ref{fig:intro} illustrates a typical instance of this task setup.

\noindent \textbf{Illustrative Example.}
As shown in Figure~\ref{fig:intro}, the story consists of four sentences, each describing an action that updates the characters’ belief state. The goal of ToM reasoning is to infer the sequence of belief states, culminating in the final belief state.

In this example, after Action 1, John, Mary, and Alice are all present in the kitchen, but the chocolate has not been introduced, so no beliefs are yet established. After Action 2, John places the chocolate in the drawer, and everyone present observes this action. Hence, Mary believes that John believes the chocolate is in the drawer. Following Action 3, John exits the kitchen. Then, in Action 4, Mary moves the chocolate to the table, an action that John is unaware of. As a result, Mary thinks John still believes the chocolate is in the drawer.

From this example, we see that ToM reasoning can be understood as an action applied to a prior belief state, causing characters to gain or lose information and thereby forming a new state. This process naturally aligns with DEL, which represents each belief state with an epistemic model, each action with an event model, and updates beliefs via the product update by combining the state with an action. Together, these elements yield a formal dynamic-logic system that derives the full belief-state trace over time.

\noindent \textbf{Our Objective: Inference-Time Scaling for Verifiable ToM in LLMs.} 
Our goal is to enable LLMs to perform ToM reasoning in an efficient and verifiable manner. 
To this end, we adopt an inference-time scaling strategy that allocates extra compute during inference to improve the reliability of reasoning. 
This approach not only enhances the reasoning capability of large models but also allows smaller models to remain deployment-efficient while achieving performance competitive with closed-source LLMs.

\section{Inference-Time Scaling for ToM}
\begin{figure*}[t]
    \centering
    \includegraphics[width=0.92\linewidth]{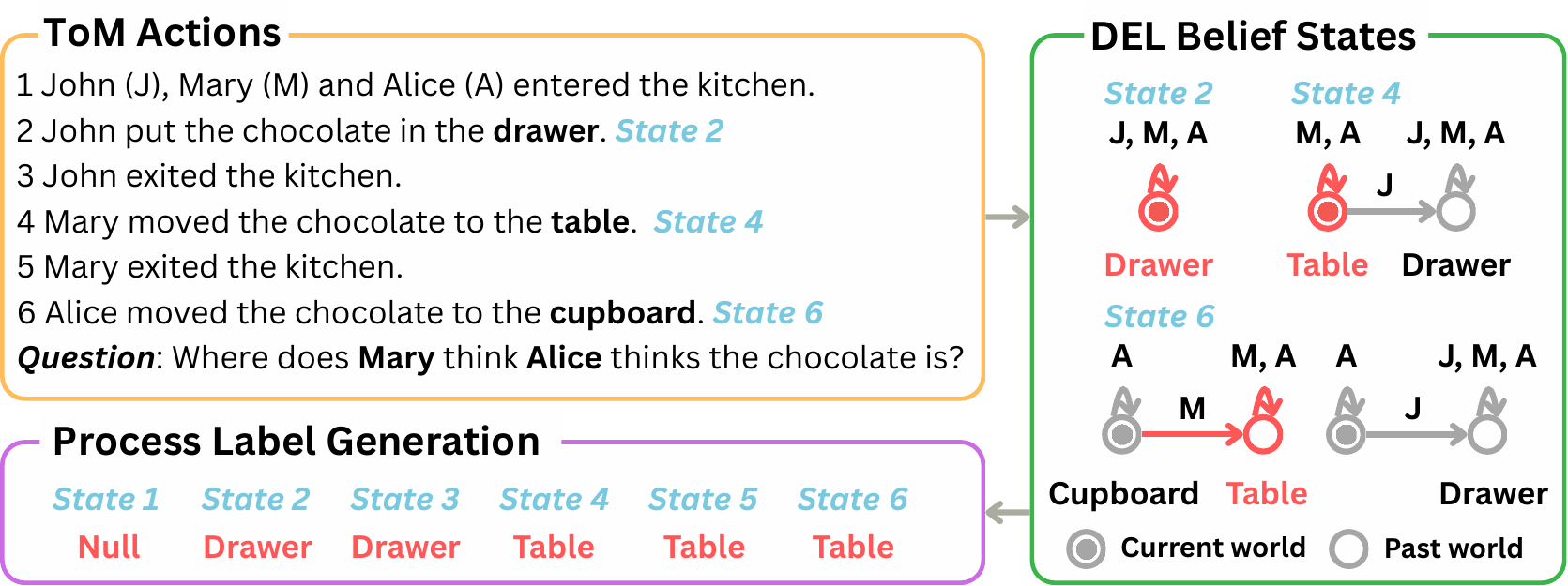}
    \caption{Training data synthesis for PBM. The right part illustrates the accessibility relations generated by the DEL simulator.}
    \label{fig:method}
\end{figure*}

In this section, we first formulate ToM reasoning as a DEL process. 
We then describe how the PBM is constructed and trained to evaluate belief traces, and present inference-time scaling pipelines that use the PBM to guide ranking and selection of reasoning traces.

\subsection{Formulating ToM Reasoning within DEL}

We formulate ToM reasoning within the framework of DEL, which is based on Kripke’s possible-world semantics~\cite{Kripke1963-KRISCO}. 
Let $\mathcal{P}$ be a countable set of atomic propositions, representing basic facts about the world, and let $\mathcal{A}$ be a finite, non-empty set of agents. 
The epistemic language $\mathcal{L}(\mathcal{P}, \mathcal{A})$ is defined by the Backus-Naur form~\cite{knuth1964backus}:
\begin{align*}
    \varphi ::= p \mid \neg \varphi \mid \varphi \wedge \varphi \mid B_i \varphi,
\end{align*}
where $p \in \mathcal{P}$, $i \in \mathcal{A}$, and $\varphi$ ranges over well-formed formulas. The formula $B_i \varphi$ is read as “agent $i$ believes $\varphi$.” 
For example, “John believes the chocolate is in the drawer” can be written as $B_{\text{John}}(\mathit{chocolate\_in\_drawer})$.
Based on this language, we define epistemic models, event models, and the product update.

\begin{definition}[Epistemic Model]
An \emph{epistemic model} over agent set $\mathcal{A}$ and proposition set $\mathcal{P}$ is a triple $\mathcal{M} = (W, R, V)$, where: 
\begin{itemize}[nosep,leftmargin=*]
    \item $W$ is a set of possible worlds, where each world is a complete valuation of $\mathcal{P}$;
 
    \item $R: \mathcal{A} \rightarrow 2^{W \times W}$ assigns each agent $a \in \mathcal{A}$ an accessibility relation $R_a$; 
    \item $V: \mathcal{P} \rightarrow 2^W$ maps each atomic proposition $p \in \mathcal{P}$ to the set of worlds where $p$ is true.
\end{itemize}
\end{definition}

A \emph{state} is a pointed epistemic model $(\mathcal{M}, w)$ where $w \in W$ is the designated actual world.

We write $w R_a v$ to denote that world $v$ is accessible from world $w$ according to agent $a$: in world $w$, agent $a$ considers $v$ possible. 

On the basis of an epistemic model $\mathcal{M} = (W,R,V)$ and a designated world $w \in W$, 
the satisfaction relation $\models$ for $\mathcal{L}(\mathcal{P}, \mathcal{A})$ is defined as follows:
\begin{itemize}[nosep,leftmargin=*]
    \item $\mathcal{M}, w \models p$ iff $w \in V(p)$;
    \item $\mathcal{M}, w \models B_a \varphi$ iff for all $v \in W$ such that $w R_a v$, we have $\mathcal{M}, v \models \varphi$.
\end{itemize}

\begin{definition}[Event Model]
An \emph{event model} is a tuple $\varepsilon = (E, Q, \mathsf{pre}, \mathsf{post})$, where:
\begin{itemize}[nosep,leftmargin=*]
    \item $E$ is a finite, non-empty set of events;
    \item $Q: \mathcal{A} \to 2^{E \times E}$ assigns to each agent $a \in \mathcal{A}$ an indistinguishability relation $Q_a$ over events;
    \item $\mathsf{pre}: E \to \mathcal{L}(\mathcal{P}, \mathcal{A})$ assigns to each $e \in E$ a precondition specifying when $e$ is executable;
    \item $\mathsf{post}: E \to \mathcal{L}(\mathcal{P}, \mathcal{A})$ assigns to each $e \in E$ a postcondition describing how the world changes.
\end{itemize}
\end{definition}

We refer to a pointed event model $(\varepsilon, e)$ as an \emph{action}, where $e \in E$ is the actual event that occurs.

\begin{definition}[Product Update]
Let $(\mathcal{M},w)$ be a state with $\mathcal{M} = (W,R,V)$, and let $(\varepsilon,e)$ be an action with $\varepsilon = (E,Q,\mathsf{pre},\mathsf{post})$. Suppose that the precondition is satisfied, i.e., $\mathcal{M}, w \models \mathsf{pre}(e)$. Then the \emph{product update} results in a new state $(\mathcal{M}', (w, e))$, where the updated epistemic model $\mathcal{M}' = (W', R', V')$ is defined as follows:
\begin{itemize}[nosep,leftmargin=*]
    \item $W' = \{(w', e') \in W \times E \mid \mathcal{M}, w' \models \mathsf{pre}(e')\}$;
    \item For each $a \in \mathcal{A}$, $R'_a = \{((w', e'), (v', f')) \in W' \times W' \mid w' R_a v' \wedge e' Q_a f'\}$;
    \item $(w', e') \in V'(p)$ iff $\mathsf{post}(e') \models p$ or $(\mathcal{M}, w' \models p \wedge \mathsf{post}(e') \not\models \neg p)$, for each $p \in \mathcal{P}$.
\end{itemize}
\end{definition}

\noindent\textbf{Applying DEL to ToM Reasoning.}
We illustrate States~4--6 in Figure~\ref{fig:method}. 
In State~4, both Mary and Alice are present and observe that the chocolate is on the table, so $\mathcal{M}, w_4 \models \mathit{table}$ and $R_M = R_A = \{(w_4, w_4)\}$. 
After Action~5, Mary exits the kitchen, $\mathsf{pre}(e_5)=\top,\ \mathsf{post}(e_5)=\mathit{table}$, so facts remain unchanged but Mary will not observe subsequent actions. In Action~6, Alice moves the chocolate to the cupboard with $\mathsf{pre}(e_6)=\top$ and $\mathsf{post}(e_6)=\mathit{cupboard}\wedge\neg\mathit{table}$. 
After the product updates, the actual state $(\mathcal{M}, w_6)$ satisfies $\mathcal{M}, w_6 \models \mathit{cupboard}$; Alice’s accessibility relation $R'_A$ points to cupboard-worlds, while Mary’s relation $R'_M$ still reaches the table-world. Hence
\begin{align*}
&\mathcal{M},w_6 \models B_{\text{Mary}}\,B_{\text{Alice}}\,\varphi \\
&\iff \forall v\,(w_6 R'_M v \Rightarrow \mathcal{M},v \models B_{\!A}\varphi)\\
&\iff \mathcal{M},w_4 \models B_{\text{Alice}}\,\varphi \\
&\iff \forall u\,(w_4 R'_A u \Rightarrow \mathcal{M},u \models \varphi)\\
&\iff \mathcal{M},w_4 \models \varphi,
\end{align*}
where $\varphi$ denotes “the chocolate is on the table.” Thus, Mary believes that Alice believes it is on the table. 
This illustrates that the core of DEL reasoning lies in constructing the \textbf{accessibility relations $R$} at each state, finding the worlds that are compatible with an agent’s belief~\cite{Hintikka1989-HINTLO-7}.

\subsection{Building the PBM with DEL}

\noindent \textbf{Generating Process-Level Labels via DEL.}
We integrate a DEL simulator into the Hi-ToM generators\footnote{\url{https://github.com/ying-hui-he/Hi-ToM_dataset}}~\cite{he2023hi} and synthesize 20{,}000 ToM stories with process labels. 
For each story, we build process-level traces across different orders of belief: at each action, we update the accessibility relations $R$ based on the action's semantics and whether the observation is public or private, then update $R$ accordingly and record the belief state in the trace set. 
All process-level label generation code is integrated into the Hi-ToM generators and included in our released codebase.

\noindent \textbf{Dataset Assembly.}
For each synthesized story, we prompt GPT-4o-mini~\cite{hurst2024gpt} to produce step-by-step belief updates in a DEL format (the prompt is provided in Appendix~\ref{app:template}). 
We pair each LLM trace with the DEL per-step labels to form training instances, yielding both positive and negative supervision for process-level reward modeling.

\noindent \textbf{Training the PBM.} PBM is a scoring function $f: \mathcal{Q} \times \mathcal{S} \to \mathbb{R}^+$ that assigns a score to each step $s_i$ in a GPT-4o-mini-generated belief trace $s$, given a ToM problem $q$. We treat this as a binary classification task: each step is labeled as either correct or incorrect according to the DEL-generated belief trace. The model is trained using the following binary cross-entropy loss:
\begin{align*}
\mathcal{L}_{\text{PBM}} 
&= - \sum_{i=1}^{K} y_{s_i} \log f(s_i) \\
&\quad - \sum_{i=1}^{K} (1 - y_{s_i}) \log (1 - f(s_i)),
\end{align*}
where $K$ is the number of steps, $y_{s_i}$ is the binary label, and $f(s_i)$ is the predicted score. The training code is adopted from the RLHF-Reward-Modeling codebase\footnote{\url{https://github.com/RLHFlow/RLHF-Reward-Modeling}}.

\subsection{Inference-Time Scaling Pipeline}

\noindent \textbf{Beam Search.} Beam search is a decoding method that maintains multiple partial belief traces during generation (Figure~\ref{fig:intro}): at each action, the LLM observes the trace so far and proposes multiple candidate belief updates for the current state. The PBM scores these candidates, and a high-scoring subset is selected to continue reasoning. 
This process repeats until all actions are processed. Formally, the procedure is as follows:
\begin{itemize}[nosep,leftmargin=*]
    \item Initialize $k$ beams with candidate first-step updates sampled from the model.
    \item Expand each beam with $b$ next-step candidates, yielding $k \times b$ partial paths.
    \item Score each path with the PBM, ranking by the score of the most recent step.
    \item Retain the top $k$ paths and iterate until reaching an end-of-sequence or the maximum depth.
\end{itemize}

\noindent \textbf{Best-of-N (BoN).} 
Alternatively, instead of updating step by step, the LLM may generate $N$ complete belief traces after reading the entire story. 
The PBM scores each step in these traces, aggregates the step-wise scores into a process-level reward, and reranks the candidates to identify the most reliable trace as the final output. 
We experiment with different aggregation rules for computing the trace-level score:
\begin{itemize}[nosep,leftmargin=*]
    \item Last: Use the PBM score of the final step.
    \item Min: Use the lowest score across all steps.
    \item Avg: Use the average score across the trace.
    \item Prod: Multiply the scores of all steps.
    \item Majority: Select the final answer by simple majority voting across traces, without using PBM.
    
\end{itemize}

Based on the aggregated scores, we consider two ranking strategies:

\begin{itemize}[nosep,leftmargin=*]
    \item Vanilla BoN: Select the single trace with the highest PBM score.
    \item Weighted BoN: Group traces by their final answers, yielding a candidate set \(\mathcal{Y} = \{y_1, y_2, \dots \}\). We then sum PBM scores within each group and select the answer $\hat{y}$ with the highest total:
    \[
    \hat{y} = \arg\max_{y \in \mathcal{Y}} \sum_{i=1}^N \mathbbm{1}(y_i = y) \cdot \mathrm{PBM}(p, t_i),
    \]
    where $t_i$ is the $i$-th trace, $y_i$ denotes the trace's final answer, and $\mathrm{PBM}(p, t_i)$ is its score.
\end{itemize}

\section{Experiments}
\begin{table*}[!h]
\centering
\small
\caption{Inference-time scaling across belief orders in the Hi-ToM dataset using BoN and Beam Search. ``Ori'' denotes baseline accuracy, and ``+PBM'' denotes accuracy with inference-time scaling.}
\label{tab:main-results}
\begin{tabular}{lcccccccccccc}
\toprule
\multirow{2}{*}{Model} & \multicolumn{2}{c}{0-th Order} & \multicolumn{2}{c}{1-th Order} & \multicolumn{2}{c}{2-th Order} & \multicolumn{2}{c}{3-th Order} & \multicolumn{2}{c}{4-th Order} & \multicolumn{2}{c}{Average} \\
\cmidrule(lr){2-3}
\cmidrule(lr){4-5}
\cmidrule(lr){6-7}
\cmidrule(lr){8-9}
\cmidrule(lr){10-11}
\cmidrule(lr){12-13}
 & Ori & +PBM & Ori & +PBM & Ori & +PBM & Ori & +PBM & Ori & +PBM & Ori & +PBM \\
\midrule
\multicolumn{13}{c}{\textit{BoN} ($N=1024$)} \\
\midrule

Qwen3-4B & 100.0 & 100.0 & 79.8 & 85.0 & 79.3 & 90.0 & 70.2 & 82.5 & 46.0 & 65.0 & 75.1 & 84.5 \\
Qwen3-1.7B & 78.0 & 82.5 & 59.7 & 65.0 & 45.2 & 55.0 & 47.0 & 62.5 & 47.8 & 57.5 & 55.5 & 64.5 \\
Qwen3-0.6B & 69.2 & 80.0 & 52.0 & 72.5 & 35.0 & 47.5 & 31.5 & 52.5 & 34.0 & 47.5 & 44.3 & 60.0 \\
Llama3.2-3B & 68.2 & 85.0 & 52.0 & 80.0 & 43.2 & 82.5 & 37.0 & 82.5 & 36.8 & 75.0 & 47.4 & 81.0 \\
Llama3.2-1B & 41.5 & 46.2 & 40.0 & 53.8 & 28.5 & 61.5 & 41.5 & 84.6 & 29.2 & 58.3 & 36.1 & 60.9\\
\midrule
\multicolumn{13}{c}{\textit{BoN} ($N=4$)} \\
\midrule
gpt-4.1        & 95.0 & 97.5 & 85.0 & 87.5 & 85.0 & 92.5 & 82.5 & 95.0 & 70.0 & 77.5 & 83.5 & 90.0 \\
gpt-4.1-mini   & 77.5 & 70.0 & 90.0 & 85.0 & 70.0 & 75.0 & 75.0 & 92.5 & 77.5 & 92.5 & 78.0 & 83.0 \\
gpt-4o         & 100.0 & 100.0 & 85.0 & 90.0 & 82.5 & 92.5 & 90.0 & 97.5 & 77.5 & 85.0 & 87.0 & 93.0 \\
gpt-4o-mini    & 90.0 & 100.0 & 75.0 & 87.5 & 77.5 & 95.0 & 77.5 & 100.0 & 55.0 & 85.0 & 75.0 & 93.5 \\
\midrule
\multicolumn{13}{c}{\textit{Beam Search} ($N=256$)} \\
\midrule
Qwen3-8B & 96.5 & 80.0 & 53.3& 80.0 &  38.8  & 85.0 & 55.8  & 95.0 & 57.8  & 95.0 &60.4  & 87.0 \\

Qwen3-4B & 100.0 & 100.0 & 79.8 & 85.0 & 79.3 & 97.5 & 70.2 & 82.5 & 46.0 & 60.0 & 75.1 & 85.0\\
\bottomrule
\end{tabular}
\end{table*}

\begin{table}[t]
\centering
\small
\caption{Comparison with SOTA LLMs on Hi-ToM (BoN, $N=1024$). ``+PBM'' denotes accuracy with inference-time scaling.}
\label{tab:sota-compare}
\resizebox{\columnwidth}{!}{
\begin{tabular}{lcccccc}
\toprule
Model & 0-th & 1-th & 2-th & 3-th & 4-th & Avg. \\
\midrule
o4-mini              & 97.5 & 95.0 & 77.5 & 87.5 & 85.0 & 88.5 \\
gpt-4o               & 100.0 & 85.0 & 82.5 & 90.0 & 77.5 & 87.0 \\
\textbf{Qwen3-4B+PBM}   & \textbf{100.0} & \textbf{85.0} & \textbf{90.0} & \textbf{82.5} & \textbf{65.0} & \textbf{84.5} \\
Qwen3-235B-A22B      & 100.0 & 75.0 & 85.0 & 85.0 & 75.0 & 84.0 \\
gpt-4.1              & 95.0 & 85.0 & 85.0 & 82.5 & 70.0 & 83.5 \\
DeepSeek-V3          & 100.0 & 80.0 & 90.0 & 70.0 & 72.5 & 82.5 \\
\textbf{Llama3.2-3B+PBM} & \textbf{85.0} & \textbf{80.0} & \textbf{82.5} & \textbf{82.5} & \textbf{75.0} & \textbf{81.0} \\
gpt-4.1-mini         & 77.5 & 90.0 & 70.0 & 75.0 & 77.5 & 78.0 \\
gpt-4o-mini          & 90.0 & 75.0 & 77.5 & 77.5 & 55.0 & 75.0 \\
\textbf{Qwen3-1.7B+PBM} & \textbf{82.5} & \textbf{65.0} & \textbf{55.0} & \textbf{62.5} & \textbf{57.5} & \textbf{64.5} \\
OLMo-32B             & 77.5 & 60.0 & 60.0 & 65.0 & 52.5 & 63.0 \\
\textbf{Llama3.2-1B+PBM} & \textbf{46.2} & \textbf{53.8} & \textbf{61.5} & \textbf{84.6} & \textbf{58.3} & \textbf{60.9} \\
\textbf{Qwen3-0.6B+PBM} & \textbf{80.0} & \textbf{72.5} & \textbf{47.5} & \textbf{52.5} & \textbf{47.5} & \textbf{60.0} \\
gpt-4.1-nano         & 22.5 & 32.5 & 42.5 & 27.5 & 30.0 & 31.0 \\
\bottomrule
\end{tabular}
}
\end{table}

\subsection{Experimental Setup}

\noindent \textbf{Platform.}
All experiments are conducted on a single NVIDIA GH200 GPU node. We use the vLLM~\cite{kwon2023efficient} framework for efficient batched inference and large-scale decoding.

\noindent \textbf{PBM Training.}
We fine-tune a PBM model based on Llama3.1-8B-Instruct~\cite{grattafiori2024llama}. The model is trained for 1 epoch using our synthesized dataset. 

\noindent \textbf{Test Models.} 
We evaluate our methods on both the Qwen3 series (0.6B, 1.7B, 4B, 8B)~\cite{yang2025qwen3} and the Llama3.2 series (1B, 3B)~\cite{grattafiori2024llama}, as well as closed-source models including gpt-4.1, gpt-4o, gpt-4.1-mini, and gpt-4o-mini. 
For comparison, we also report results from baselines such as o4-mini, gpt-4.1-nano, Qwen3-235B-A22B~\cite{yang2025qwen3}, DeepSeek-V3~\cite{liu2024deepseek}, and OLMo-2-0325-32B~\cite{walsh2}. 
All models are evaluated under their default generation settings.

\noindent \textbf{Datasets.}
We conduct evaluations on two datasets: Hi-ToM~\cite{he2023hi} and the ToM tasks introduced by Kosinski~\cite{kosinski2023theory}. 
For Hi-ToM, we only evaluate one-chapter stories, and for Kosinski’s dataset we restrict evaluation to the unexpected transfer task.

\noindent \textbf{Metrics and Prompt Format.}
We report final answer accuracy as the main evaluation metric. All models are evaluated using a consistent prompting format, as detailed in Appendix \ref{app:template}.

\subsection{Results on Hi-ToM Dataset}
For BoN, we scale $N$ up to 1024 and apply the weighted strategy, selecting the best aggregation rule for each instance. For beam search, we evaluate Qwen3-4B and Qwen3-8B with beam sizes from 4 to 256, excluding smaller models since they cannot generate valid intermediate reasoning steps.

\noindent \textbf{Main Results.} 
As shown in Table~\ref{tab:main-results}, incorporating PBM consistently improves ToM reasoning across both BoN and beam search. 
For example, Llama3.2-3B gains 33.6 points in average accuracy, while Qwen3-4B improves by 9.4 points in the BoN setting. 
Similarly, with beam search, Qwen3-8B, whose baseline underperforms Qwen3-4B, achieves the highest accuracy of 87.0 once guided by PBM. 
Moreover, our method generalizes to both open- and closed-source models, as the gpt series also shows clear gains with PBM.

\noindent \textbf{Comparison with SOTA LLMs.}  
As shown in Table~\ref{tab:sota-compare}, smaller open-source models can match or surpass much larger LLMs. For example, Qwen3-4B+PBM achieves higher average accuracy than gpt-4.1, DeepSeek-V3, and OLMo-32B, while Llama3.2-3B+PBM performs on par with gpt-4.1-mini. These findings highlight the effectiveness of PBM in scaling ToM reasoning.

\begin{figure*}[t]
    \centering
    \begin{subfigure}[t]{0.48\textwidth}
        \centering
        \includegraphics[width=\textwidth]{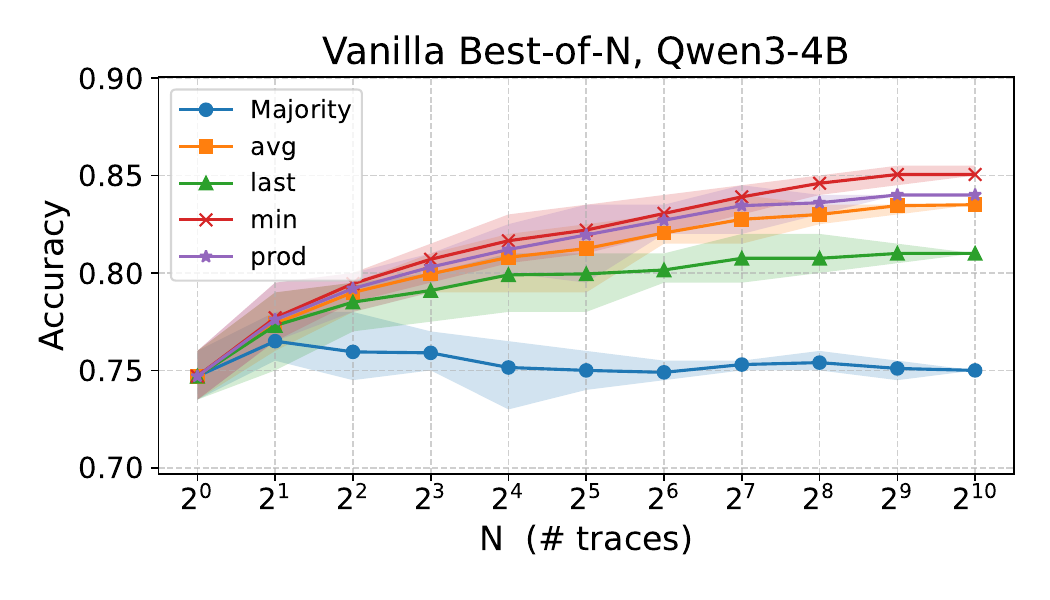}
        \caption{Vanilla BoN decoding on Qwen3-4B.}
    \end{subfigure}
    \hfill
    \begin{subfigure}[t]{0.48\textwidth}
        \centering
        \includegraphics[width=\textwidth]{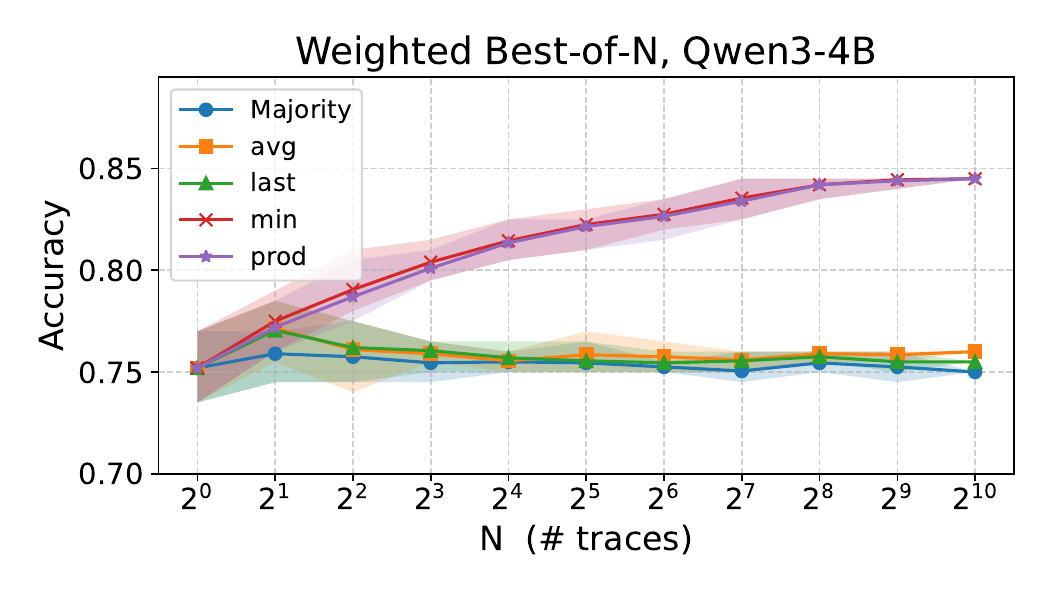}
        \caption{Weighted BoN decoding on Qwen3-4B.}
    \end{subfigure}
    \caption{Accuracy of BoN decoding on Qwen3-4B across different budgets $N$ in the Hi-ToM dataset. Results are shown for (a) Vanilla and (b) Weighted aggregation strategies.}
    \label{fig:bon-curve}
\end{figure*}

\noindent \textbf{Scaling Test-Time Compute for ToM Reasoning.}  
As shown in Figure~\ref{fig:bon-curve}, increasing the number of sampled belief traces $N$ improves ToM performance only when guided by PBM. Among aggregation strategies, min and prod are the most reliable, while avg and last often degrade under weighted aggregation. In contrast, majority voting fails to improve accuracy, since ToM requires evaluating intermediate belief states rather than aggregating final answers. A theoretical analysis of this limitation is provided in Appendix~\ref{app:proof}.

\noindent \textbf{BoN vs.\ Beam Search.}
Our experiments show that these two inference-time strategies achieve comparable accuracy. However, beam search rollouts often fail on smaller or weaker models that cannot reliably produce valid intermediate states, making PBM evaluation infeasible. In contrast, BoN generates full belief traces in one shot, where PBM remains effective even when some steps are noisy, and large candidate sets can be produced efficiently using high-throughput backends such as vLLM. We therefore recommend BoN as the preferred inference-time scaling method for ToM reasoning.

\subsection{Results on Out-of-Distribution ToM Data}

Our PBM is trained on Hi-ToM-style synthetic data, but we ask: \emph{Can it generalize to ToM tasks from a different distribution?} 
To test this, we evaluate it on the dataset from Kosinski~\cite{kosinski2023theory}, which contains hand-written scenarios with false-belief and true-belief controls. 
We experiment with the Qwen3 series, following the same inference-time scaling and PBM-based selection procedure as before.

\noindent \textbf{Main Results.}  
As shown in Table~\ref{tab:kosinski-results}, PBM improves accuracy across all models, confirming its ability to generalize beyond synthetic Hi-ToM scenarios. 
This shows that PBM functions as a genuine verifier of whether a ToM reasoning process is justified, rather than overfitting to the training distribution, and highlights its robustness on out-of-domain ToM tasks.

\begin{table*}[t]
\centering
\small
\caption{BoN ($N=1024$) inference-time scaling on the dataset from Kosinski~\cite{kosinski2023theory}, evaluated across different belief types. ``Ori'' denotes baseline accuracy; ``+PBM'' denotes accuracy with inference-time scaling.}
\label{tab:kosinski-results}
\begin{tabular}{lcccccccccc}
\toprule
\multirow{2}{*}{Model} & \multicolumn{2}{c}{False Belief} & \multicolumn{2}{c}{Informed Protagonist} & \multicolumn{2}{c}{No Transfer} & \multicolumn{2}{c}{Present Protagonist} & \multicolumn{2}{c}{Average} \\
\cmidrule(lr){2-3}
\cmidrule(lr){4-5}
\cmidrule(lr){6-7}
\cmidrule(lr){8-9}
\cmidrule(lr){10-11}
 & Ori & +PBM & Ori & +PBM & Ori & +PBM & Ori & +PBM & Ori & +PBM \\
\midrule
Qwen3-8B & 83.3 & 87.5 & 83.8 & 85.0 & 92.8 & 97.5 & 79.5 & 85.0 & 84.8 & 88.8 \\
Qwen3-4B & 70.2 & 80.0 & 86.2 & 90.0 & 93.2 & 95.0 & 88.0 & 92.5 & 84.4 & 89.4 \\
Qwen3-1.7B & 18.2 & 35.0 & 15.5 & 37.5 & 24.8 & 60.0 & 13.8 & 30.0 & 18.1 & 40.6 \\
Qwen3-0.6B & 14.5 & 12.5 & 23.5 & 30.0 & 25.0 & 35.0 & 21.0 & 32.5 & 21.0 & 27.5 \\

\bottomrule
\end{tabular}
\end{table*}

\subsection{Benchmarking the PBM}

To assess the PBM’s standalone reliability, we construct a held-out test set of 2,000 multi-step reasoning examples generated by gpt-4o-mini and spanning all belief orders. 
Each step is labeled by the DEL simulator as either correct or incorrect, and the PBM is evaluated by its step-level classification accuracy. We benchmark two PBMs trained on different base models: Llama3.1-8B-Instruct and Llama3.2-3B-Instruct.

\begin{table}[h]
\centering
\small
\caption{PBM classification accuracy (\%) across belief orders on the test set.}
\label{tab:pbm-classify}
\resizebox{\columnwidth}{!}{
\begin{tabular}{lcccccc}
\toprule
PBM & 0-th & 1-th & 2-th & 3-th & 4-th & Avg. \\
\midrule
Llama3.1-8B & 99.2 & 94.6 & 89.0 & 87.0 & 79.9 & 90.0 \\
Llama3.2-3B & 99.1 & 91.9 & 84.9 & 83.8 & 73.8 & 86.7 \\
\bottomrule
\end{tabular}
}
\end{table}

\begin{table}[h]
\centering
\small
\caption{BoN inference-time scaling accuracy (\%) on Hi-ToM using different PBMs.}
\label{tab:pbm-downstream}
\resizebox{\columnwidth}{!}{
\begin{tabular}{lcccccc}
\toprule
Model+PBM & 0-th & 1-th & 2-th & 3-th & 4-th & Avg. \\
\midrule
Qwen3-4B + 8B  & 100.0 & 85.0 & 90.0 & 82.5 & 65.0 & 84.5 \\
Qwen3-4B + 3B  & 100.0 & 77.5 & 77.5 & 72.5 & 47.5 & 75.0 \\
Qwen3-1.7B + 8B & 82.5 & 65.0 & 55.0 & 62.5 & 57.5 & 64.5 \\
Qwen3-1.7B + 3B & 82.5 & 60.0 & 45.0 & 47.5 & 50.0 & 57.0 \\
Qwen3-0.6B + 8B & 80.0 & 72.5 & 47.5 & 52.5 & 47.5 & 60.0 \\
Qwen3-0.6B + 3B & 77.5 & 55.0 & 27.5 & 35.0 & 32.5 & 45.5 \\
\bottomrule
\end{tabular}
}
\end{table}

\noindent \textbf{Evaluating PBM.} As shown in Table~\ref{tab:pbm-classify}, the larger PBM achieves consistently higher accuracy, and performance decreases as the belief order increases. This suggests that stronger models can better verify reasoning steps, while evaluating deeper recursive beliefs is inherently more challenging.

\noindent \textbf{Impact of PBM Quality on Task Accuracy.}  We further test how the quality of the PBM affects end-task performance. 
Specifically, we run BoN inference-time scaling on the Hi-ToM dataset using different base models, guided either by a strong PBM (Llama3.1-8B-Instruct) or a weaker one (Llama3.2-3B-Instruct).
As shown in Table~\ref{tab:pbm-downstream}, replacing the strong PBM with a weaker one consistently reduces accuracy across all base models and belief orders. 
This establishes a clear link between verifier quality and final task performance: a stronger PBM leads to better inference-time scaling outcomes.

\noindent \textbf{Qualitative Analysis of PBM Behavior.}  
To better understand when PBM succeeds or fails, we examine its behavior on reasoning traces. 
Below are two steps predicted by the Llama3.2-3B-Instruct PBM.

Scenario:
Initially, everyone knows that the asparagus is in the \texttt{blue\_cupboard}. 
At the current moment, Charlotte and Elizabeth are present in the room, while Alexander has just left.
Charlotte holds a second-order belief about Alexander’s belief regarding Elizabeth.

\noindent{Step $n$:}  
\begin{itemize}[nosep,leftmargin=*]
    \item \textbf{Action:} \texttt{Elizabeth likes the red\_box.}  
    \item \textbf{State:} Irrelevant. Charlotte thinks Alexander thinks Elizabeth thinks the asparagus is in \texttt{blue\_cupboard}.  
    \item \textbf{Prediction:} $+$ \quad \textbf{Ground Truth:} $+$  
    \item \textbf{Annotation:} This step is correct. The statement is unrelated to the asparagus; no beliefs update. PBM correctly captures this invariance.
\end{itemize}

\noindent{Step $n\!+\!1$:}  
\begin{itemize}[nosep,leftmargin=*]
    \item \textbf{Action:} \texttt{Elizabeth moved the asparagus to the green\_bucket.}  
    \item \textbf{State:} Only Elizabeth and Charlotte are present when this happens. Charlotte sees this move. Charlotte thinks Alexander thinks Elizabeth thinks the asparagus is in \texttt{green\_bucket}.  
    \item \textbf{Prediction:} $+$ \quad \textbf{Ground Truth:} $-$  
    \item \textbf{Annotation:} This step is incorrect. Since Alexander is not present, he cannot observe Elizabeth’s action. Therefore, his beliefs (as perceived by Charlotte) should not change. PBM overgeneralizes belief update based on partial presence.
\end{itemize}

This example shows that while PBM handles simple irrelevant statements, it can fail on nested, perspective-sensitive updates, revealing a key challenge in verifying multi-agent reasoning.

\subsection{Discussion}

\noindent \textbf{Cost Efficiency for API-based Usage.}
As shown in Table~\ref{tab:main-results}, applying PBM narrows the gap between small and large models: gpt-4.1-mini approaches gpt-4.1, while gpt-4o-mini gains +18.5 points, surpassing gpt-4o. 
Despite sampling $N{=}4$ outputs, mini models remain more cost-efficient, with per-million-token costs of only 2.10 and 0.825 compared to 10.50 and 13.75 for the larger models (Table~\ref{tab:api-cost}). 
Furthermore, because all $N$ samples share the same input prompt, the input cost is paid only once, and only the output tokens scale with $N$. This makes PBM-guided small-batch inference-time scaling a cheaper alternative to using larger models.

\begin{table}[t]
\centering
\small
\caption{API price per 1M tokens.}
\label{tab:api-cost}
\resizebox{\columnwidth}{!}{
\begin{tabular}{lcccc}
\toprule
Model & Input & Cached Input & Output & Total \\
\midrule
gpt-4.1      & \$2.00 & \$0.50  & \$8.00  & \$10.50 \\
gpt-4.1-mini & \$0.40 & \$0.10  & \$1.60  & \$2.10  \\
gpt-4o       & \$2.50 & \$1.25  & \$10.00 & \$13.75 \\
gpt-4o-mini  & \$0.15 & \$0.075 & \$0.60  & \$0.825 \\
\bottomrule
\end{tabular}}
\end{table}

\begin{figure}[t]
    \centering
    \includegraphics[width=0.485\textwidth]{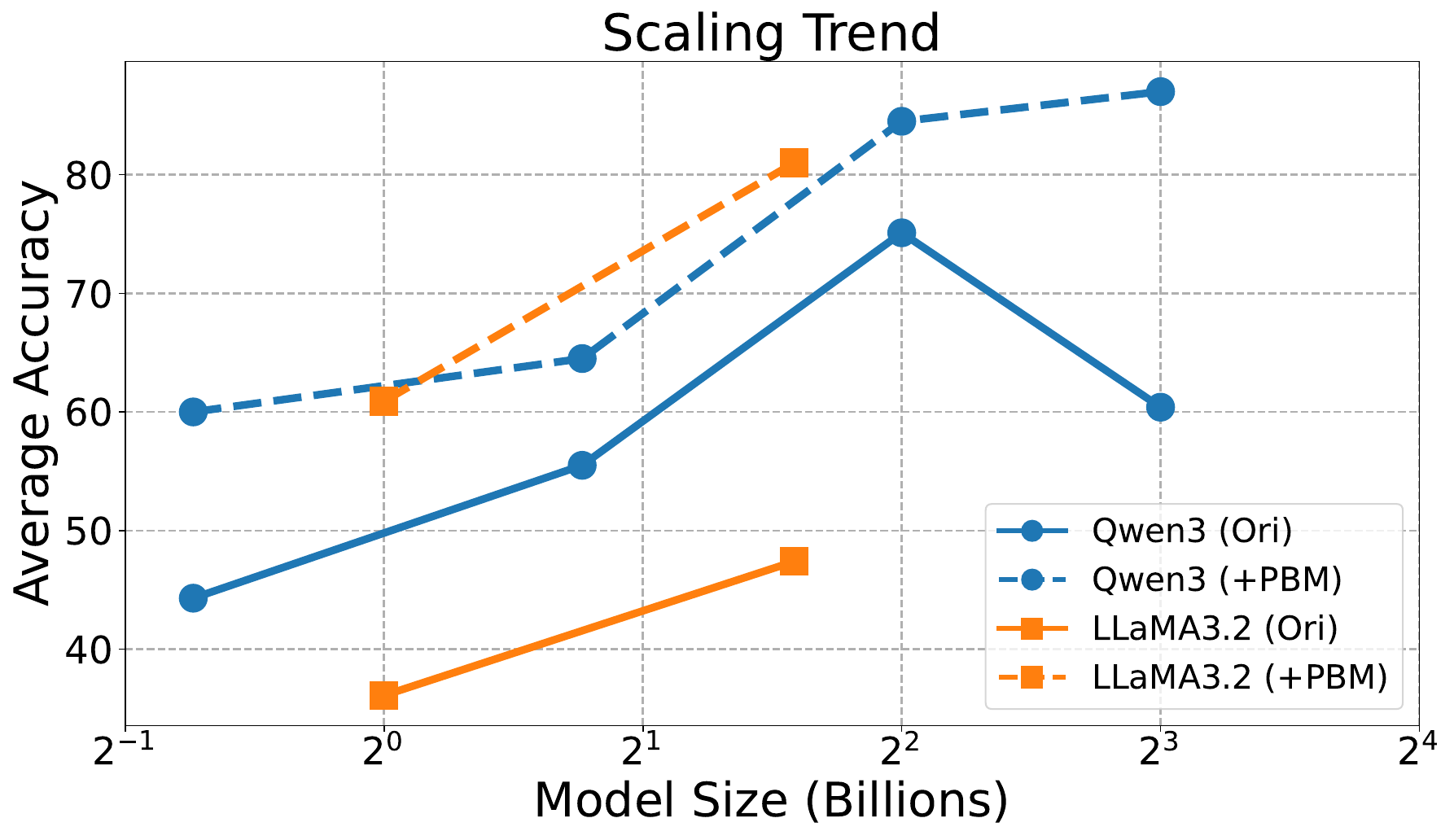}
    \caption{Scaling trend of average accuracy before and after applying PBM across different LLMs on Hi-ToM. ``Ori'' denotes baseline accuracy; ``+PBM'' denotes accuracy with inference-time scaling.}
    \label{fig:scaling-trend}
\end{figure}

\noindent \textbf{Scaling with Model Size.}
Figure~\ref{fig:scaling-trend} shows how ToM accuracy changes with model size. PBM consistently improves performance and strengthens the scaling trend. For Llama 3.2, the accuracy curve becomes steeper when equipped with PBM, suggesting that larger models benefit more and generalize better under our inference-time intervention. Interestingly, Qwen3-8B performs worse than Qwen3-4B under the vanilla setting, but becomes the best-performing variant once PBM is applied. This indicates that PBM not only boosts accuracy but can also unlock higher-order reasoning abilities that remain latent in the base model.

\noindent \textbf{Comparison with RL-based Methods.}
Recent work~\cite{lu2025theory} has explored fine-tuning LLMs with ToM supervision using GRPO~\cite{shao2024deepseekmath} to enhance their ToM abilities. However, GRPO requires substantial compute and is notoriously difficult to optimize. In contrast, our PBM is lightweight and efficient: it trains in under three hours on a single GH200 GPU and can be applied to any target model without retraining. GRPO must be re-trained for each model and may even degrade performance on unrelated tasks such as GSM8K~\cite{lu2025theory}. Our method avoids this issue entirely by leaving model parameters unchanged. PBM thus offers a practical, generalizable, and non-invasive alternative for improving ToM reasoning.

\section{Related Work}
\noindent \textbf{DEL and Its Connections to ToM.}
DEL builds on a line of work in epistemic logic, tracing back to Hintikka’s possible-world model of knowledge and belief~\cite{Hintikka1962-HINKAB-3} and Kripke’s formal semantics~\cite{Kripke1963-KRISCO}, and later evolving through studies on information change~\cite{baltag98logic}.
It was later formalized as a unified framework of epistemic and event models with product updates~\cite{van2007dynamic} for representing and updating agents’ beliefs.
This aligns naturally with the core of ToM, which concerns reasoning about others’ beliefs. Early cognitive models used DEL to simulate belief change in multi-agent settings~\cite{bolander2011epistemic}, showing its suitability for structured belief reasoning. More recent work uses logic-based simulators to supply symbolic supervision for belief updates~\cite{bolander2014seeing,hansen2020implementing}. Building on this line, we use DEL not only as a formalism for modeling beliefs but also as a scaffold for inference-time scaling, enabling compositional and verifiable reasoning in ToM tasks.

\noindent \textbf{Inference-Time Scaling of LLMs.}  
Recent work investigates inference-time scaling as an alternative to increasing model size for improving reasoning capabilities~\cite{beeching2024scalingtesttimecompute,muennighoff2025s1}.  
Two main paradigms have been studied.  
One is single-trace scaling, which encourages deeper reasoning within a single inference path, often via reinforcement learning~\cite{guo2025deepseek,cheng2025reasoning} or distillation from a stronger teacher~\cite{li_2025_llms}.  
The other is multi-trace scaling~\cite{brown_2024_large, snell2025scaling,  schaeffer2025large}, which generates multiple reasoning traces in parallel and selects the best outcome using voting~\cite{wangself,wang_2025_ranked} or external verifiers~\cite{wang2024math,sun2024easy,guo_2025_reward,saadfalcon_2025_shrinking}.  
Recent work further combines multi-trace generation with search algorithms such as tree search and beam search to refine reasoning step by step~\cite{zhang2024rest,lin2025leveraging}.  
Our approach follows the multi-trace paradigm and introduces PBM-guided selection, extending inference-time scaling to ToM tasks.

\section{Conclusion}
This work introduces DEL-ToM, a framework that enhances Theory-of-Mind (ToM) reasoning in LLMs through inference-time scaling. By modeling belief updates with Dynamic Epistemic Logic (DEL) and training a verifier using DEL-generated labels, our approach enables structured and verifiable dynamic logical reasoning. DEL-ToM improves ToM performance across models and datasets, demonstrating that logical reasoning can be strengthened through formal logic and inference-time supervision. This opens new avenues for deploying ToM-capable LLMs in resource-constrained settings without retraining.

\bibliography{custom}

\newpage
\appendix

\twocolumn

\section*{Appendix}
\label{sec:appendix}
\section{Prompt Templates}
\label{app:template}
The full prompt templates are shown in Figures~\ref{fig:inf-prompt-1}, \ref{fig:inf-prompt-2}, and \ref{fig:inf-prompt-3}.

\begin{figure*}[t] 
\centering
\begin{tcolorbox}[title=One-Shot Prompt - Part 1,
  colback=gray!5, colframe=gray!50,
  boxrule=0.8pt, arc=4pt, outer arc=4pt,
  width=\textwidth]  

\begin{lstlisting}[basicstyle=\ttfamily\small, breaklines=true, mathescape=true]
Here is a story that unfolds in chronological order.

You will be asked a question about the story, which may involve either:
(1) Locating an object, or
(2) Inferring an agent's mental state (e.g., what A thinks B thinks C thinks).

To solve it, think step-by-step. At each step, repeat the current line from the story, then explain its effect on beliefs. Use [Null] if someone does not yet have knowledge. If a belief chain cannot be formed (e.g., some agent exited too early), freeze belief at the last available step.

<Note>
{note}

In public or private communication:
- The speaker believes the listener will believe the claim.
- If the listener exited the room earlier than the speaker, they will believe it.

If the question is zero-order (e.g., "Where is X really?"), then in each step, only track the actual location of the object (e.g., "X is in [Y]"). You do not need to track nested beliefs.

Here is an example:
<Story>
1 Amelia, Chloe, Liam, Owen and Benjamin entered the TV_room.
2 The celery is in the red_envelope.
3 Amelia made no movements and stayed in the TV_room for 1 minute.
4 Chloe lost his watch.
5 Amelia exited the TV_room.
6 Chloe moved the celery to the green_bucket.
7 Chloe exited the TV_room.
8 Liam moved the celery to the red_bathtub.
9 Liam exited the TV_room.
10 Owen made no movements and stayed in the TV_room for 1 minute.
11 Owen exited the TV_room.
12 Benjamin made no movements and stayed in the TV_room for 1 minute.
13 Benjamin exited the TV_room.
14 Amelia, Chloe, Liam, Owen and Benjamin entered the waiting_room.
15 Liam publicly claimed that celery is in the white_bathtub now.
16 Benjamin privately told Liam that the celery is in the blue_drawer now.

<Question>
Where does Owen think Liam thinks Chloe thinks the celery is?

<Trace>
## Step 1 ##
Amelia, Chloe, Liam, Owen and Benjamin entered the TV_room.  
Everyone is present, but the celery's location is still unknown.  
Owen thinks Liam thinks Chloe thinks the celery is in [Null]

## Step 2 ##
The celery is in the red_envelope.  
Everyone observes this.  
Owen thinks Liam thinks Chloe thinks the celery is in [red_envelope]

## Step 3 ##
Amelia made no movements and stayed in the TV_room for 1 minute.  
No effect.  
Owen thinks Liam thinks Chloe thinks the celery is in [red_envelope]
\end{lstlisting}

\end{tcolorbox}
\caption{One-Shot Prompt - Part 1.}
\label{fig:inf-prompt-1}
\end{figure*}

\begin{figure*}[t] 
\centering
\begin{tcolorbox}[title=One-Shot Prompt - Part 2,
  colback=gray!5, colframe=gray!50,
  boxrule=0.8pt, arc=4pt, outer arc=4pt,
  width=\textwidth]  

\begin{lstlisting}[basicstyle=\ttfamily\small, breaklines=true, mathescape=true]
## Step 4 ##
Chloe lost his watch.  
Irrelevant.  
Owen thinks Liam thinks Chloe thinks the celery is in [red_envelope]

## Step 5 ##
Amelia exited the TV_room.  
Irrelevant.  
Owen thinks Liam thinks Chloe thinks the celery is in [red_envelope]

## Step 6 ##
Chloe moved the celery to the green_bucket.  
Only Chloe, Liam, Owen, Benjamin are present. They all see this move.  
Owen thinks Liam thinks Chloe thinks the celery is in [green_bucket]

## Step 7 ##
Chloe exited the TV_room.  
Chloe's belief frozen; still [green_bucket]  
Owen thinks Liam thinks Chloe thinks the celery is in [green_bucket]

## Step 8 ##
Liam moved the celery to the red_bathtub.  
Only Liam, Owen, Benjamin present. They observe the move. Chloe not present, so her belief unchanged.  
Owen thinks Liam thinks Chloe thinks the celery is in [green_bucket]

## Step 9 ##
Liam exited the TV_room.  
No change.  
Owen thinks Liam thinks Chloe thinks the celery is in [green_bucket]

## Step 10 ##
Owen made no movements and stayed in the TV_room for 1 minute.  
Irrelevant. 
Owen thinks Liam thinks Chloe thinks the celery is in [green_bucket]

## Step 11 ##
Owen exited the TV_room.  
Owen's belief frozen.  
Owen thinks Liam thinks Chloe thinks the celery is in [green_bucket]

## Step 12 ##
Benjamin made no movements and stayed in the TV_room for 1 minute.  
Irrelevant.  
Owen thinks Liam thinks Chloe thinks the celery is in [green_bucket]

## Step 13 ##
Benjamin exited the TV_room.  
No change.  
Owen thinks Liam thinks Chloe thinks the celery is in [green_bucket]

## Step 14 ##
Everyone entered the waiting_room.  
No effect on beliefs.  
Owen thinks Liam thinks Chloe thinks the celery is in [green_bucket]
\end{lstlisting}

\end{tcolorbox}
\caption{One-Shot Prompt - Part 2.}
\label{fig:inf-prompt-2}
\end{figure*}

\begin{figure*}[t] 
\centering
\begin{tcolorbox}[title=One-Shot Prompt - Part 3,
  colback=gray!5, colframe=gray!50,
  boxrule=0.8pt, arc=4pt, outer arc=4pt,
  width=\textwidth]  

\begin{lstlisting}[basicstyle=\ttfamily\small, breaklines=true, mathescape=true]
## Step 15 ##
Liam publicly claimed that celery is in the white_bathtub now.  
Owen hears this statement. However, public speech only affects first- and second-order beliefs (e.g., what Liam believes, what Owen thinks Liam believes, and what Liam thinks Owen believes). It does not change Owen's belief about what Liam thinks Chloe thinks.  
Owen thinks Liam thinks Chloe thinks the celery is in [green_bucket]

## Step 16 ##
Benjamin privately told Liam that the celery is in the blue_drawer now.  
Owen does not hear this, but more importantly, private communication only affects beliefs between the speaker and the listener. It can change what Liam believes (based on exit order), or what Liam thinks Benjamin believes (based on exit order), or what Benjamin thinks Liam believes (always change) - but it cannot affect higher-order beliefs. So this does not change Owen's belief about what Liam thinks Chloe thinks.

Owen thinks Liam thinks Chloe thinks the celery is in [green_bucket]

Final Answer: [green_bucket]

Now it's your turn.

<Story>
{story}

<Question>
{question}

Give a step-by-step trace as in the example. Then, give the final answer in one line like:  
Final Answer: [your choice]

<trace>
\end{lstlisting}

\end{tcolorbox}
\caption{One-Shot Prompt - Part 3.}
\label{fig:inf-prompt-3}
\end{figure*}

\section{What Makes PBM Different from Majority Voting?}
\label{app:proof}

Here, we provide a analysis to compare PBM with majority voting.

\noindent \textbf{Problem Setup and Notation.}
Given an input $x$, the language model $\pi_\theta$ must perform $K$ sequential belief-updating steps, generating a trajectory
\[
(z_1, z_2, \dots, z_K), \quad z_i \sim \pi_\theta(\cdot \mid x, z_{< i}).
\]
After the trajectory is complete, it outputs a final answer $y$, chosen from a set of $L$ candidates (typically $L \approx 5\!-\!6$ in HiToM). We assume:
\begin{itemize}[nosep,leftmargin=*]
    \item Each step is independently correct with probability $q$.
    \item A trajectory is \emph{Good} if all $K$ steps are correct: $\Pr[\text{Good}] = q^K$.
    \item Otherwise, it is \emph{Bad}: $\Pr[\text{Bad}] = 1 - q^K$.
\end{itemize}

\noindent\textbf{Majority Voting.}
We sample $N$ i.i.d. trajectories and return the most frequent final answer.
Let:
\begin{itemize}[nosep,leftmargin=*]
    \item $G \sim \text{Binomial}(N, q^K)$: number of Good trajectories.
    \item $R = N - G$: number of Bad trajectories.
\end{itemize}
Under the uniform scattering assumption, Bad votes are evenly spread over $L - 1$ wrong answers:
\begin{align*}
    B_j \mid R \sim \text{Binomial}\!\left(R, \frac{1 - q^K}{L - 1} \right), \\ j = 1, \dots, L - 1.
\end{align*}
Majority voting succeeds iff
\[
E_{\text{maj}} = \{ G \ge 1 \ \text{and}\ G > \max_j B_j \}.
\]

\noindent\textbf{PBM Reranking.}
We sample $N$ trajectories and assign each a stepwise score:
\[
s(z) = \frac{1}{K} \sum_{k=1}^{K} \mathbf{1}\{z_k \text{ correct}\}.
\]
Good trajectories receive score $1$, and Bad trajectories score at most $1 - \frac{1}{K}$.
We select the trajectory with the highest score. PBM thus succeeds iff
\[
E_{\text{pbm}} = \{ G \geq 1 \}.
\]

\noindent\textbf{PBM Success Rate.}
By independence:
\[
A_{\text{pbm}} = \Pr(E_{\text{pbm}}) = 1 - (1 - q^K)^N.
\]

\noindent\textbf{Majority Voting Success Rate.}
We have
\begin{align*}
    A_{\text{maj}}&=\Pr(E_{\text{maj}})\le \Pr(G\ge 1)\\&=1-(1-q^K)^N = A_{\text{pbm}}.
\end{align*}
Hence, PBM always outperforms majority voting.

\noindent\textbf{Remarks.}
For majority voting, divide both sides of $G > \max_j B_j$ by $N$ and take the limit as $N \to \infty$. By the law of large numbers:
\[
\frac{G}{N} \to q^K, \qquad 
\frac{\max_j B_j}{N} \to \frac{1 - q^K}{L - 1}.
\]
A necessary condition for success is therefore:
\[
q^K > \frac{1 - q^K}{L - 1} \quad \Leftrightarrow \quad q^K > \frac{1}{L}.
\]
If $q^K \le \frac{1}{L}$ (typical for small models or hard ToM question), then
\[
\lim_{N \to \infty} A_{\text{maj}} = 0.
\]
In conclusion, majority voting is vulnerable to \emph{vote dilution}: if $q^K \le \frac{1}{L}$, Bad trajectories cluster on wrong answers and can dominate.

This analysis explains why PBM offers more reliable inference than majority voting, especially in complex ToM settings.

\end{document}